\documentclass[10pt, a4paper]{article}

\usepackage[final]{lrec2026} 
\usepackage{array} 
\usepackage{todonotes}

\title{Large Language Models for Citation Function Classification}

\name{Daniel Vodi\v{c}ka\textsuperscript{*}, Jakub \v{S}m\'{i}d\textsuperscript{*, $\dagger$},
Pavel Kr\'{a}l\textsuperscript{*,$\dagger$},
Christophe Cerisara\textsuperscript{$\ddagger$}
} 

\address{\textsuperscript{*}Department of Computer Science and Engineering\\
         \textsuperscript{$\dagger$}NTIS -- New Technologies for the Information Society\\
  University of West Bohemia in Pilsen, Faculty of Applied Sciences \\
          Univerzitní 2732/8, 301 00 Pilsen, Czech Republic \\
          \textsuperscript{$\ddagger$}CNRS LORIA, Universit\'e de Lorraine,
          Nancy, France\\
         \texttt{\{vodickad, jaksmid, pkral\}@kiv.zcu.cz, cerisara@loria.fr}\\
         \url{https://nlp.kiv.zcu.cz}}

\abstract{
Citation function classification plays a crucial role in understanding the relationships between scientific publications and advancing bibliometric analysis.  This study presents one of the first comprehensive evaluations of multiple state-of-the-art (SOTA) large language models (LLMs) for citation function classification, achieving new SOTA results on the ACL-ARC dataset. We systematically compare five models (Mistral 7B, Orca 2-7B, LLaMA 3.1-8B, Falcon 7B, and SciBERT) across zero-shot, few-shot, and fine-tuning approaches. Our fine-tuned Falcon 7B model achieves a 73,3\% macro F1 score on ACL-ARC,  representing a significant improvement over previous methods. 
Additionally, we introduce AC$^3$, a novel dataset featuring a seven-category annotation scheme that distinguishes between neutral acknowledgments and explicit evaluative stances (more opinion-oriented citations -- criticizing, complimenting, contradicting). The dataset is implemented across four context extraction variants to systematically evaluate the impact of contextual scope on classification performance.
We also provide detailed analysis of model performance, experimental configurations, and limitations to guide future research in this domain.
To our knowledge, this is one of the first studies dedicated to comprehensive model comparison for citation function classification, addressing a gap identified in recent surveys. 
 \\ \newline  }

\begin{document}
\maketitleabstract


\section{Introduction}
The exponential growth of scientific literature presents unprecedented challenges for researchers navigating the vast landscape of academic publications. Citations serve as fundamental connectors between scientific works, yet their significance extends far beyond simple reference counting \citep{6bc982f19b0a4889993a9f34bee26ba8, HASSAN2020105383}. Understanding how and why publications cite each other provides critical insights into knowledge development, research impact assessment, and the evolution of scientific discourse.

Traditional citation analysis treats all citations equally, failing to capture the nuanced relationships they represent. A citation providing supporting evidence carries different weight than one presenting contradictory findings or merely mentioning related work. This limitation has motivated the development of citation function classification systems that automatically categorize citations based on their purpose and context.

Recent advances in natural language processing (NLP), particularly large language models (LLMs), offer new possibilities for citation analysis. These models can capture semantic relationships and contextual information that traditional feature-based approaches often miss. However, systematic evaluation of modern LLMs for citation function classification remains limited, with most studies focusing on individual models or older architectures \citep{ZHANG2025101608}.

This work addresses three key contributions to the field. First, we achieve new SOTA results on the widely-used ACL-ARC dataset \citep{jurgens2016citationclassificationbehavioralanalysis}, with our best model (fine-tuned Falcon 7B) reaching 73,3\% macro F1 score, substantially outperforming previous approaches \citep{cohan2019structuralscaffoldscitationintent, 10.1145/3583780.3615018, nambanoor-kunnath-etal-2022-dynamic, 10.1109/JCDL57899.2023.00017}. Second, we present the first comprehensive comparison of multiple modern LLMs across different methods for citation function classification. As noted by \citet{ZHANG2025101608}, no previous survey has undertaken such systematic model comparison, representing a significant gap in understanding optimal approaches for this task.

Beyond addressing the comparative evaluation gap, this work introduces a novel corpus to distinguish citations with different meanings. For example, citations that criticize work, compliment it, or present contradictory findings are typically classified under the same category despite their distinct implications. To address this, our dataset with a seven-category scheme distinguishing neutral from opinion-oriented citations.
This dataset is freely available for research purposes\footnote{https://github.com/DanielRafaello/ArXiv-Citation-Classification-Corpus}, representing an additional contribution of this paper.

Our experimental evaluation encompasses five models across three learning paradigms: zero-shot prompting, few-shot prompting, and fine-tuning. 

\section{Related Work}
This section reviews previous research related to citation classification, including commonly used classification schemes.

\subsection{Citation Classification Schemes and Datasets}
\label{subsec:citation_classification_schemes_and_datasets}
The systematic classification of citation functions has evolved significantly since \citet{garfield1965can} pioneering work in 1965, which was one of the first studies, where author observed motivations for citing. \citet{teufel2006automatic} advanced this field by introducing a corpus of 2,829 annotated citations across 12 distinct classes, establishing one of the first substantial datasets for automatic citation classification.

After that, influential work proposed a six-category classification scheme that has become widely adopted in subsequent research \citep{10.1162/tacl_a_00028, jurgens2016citationclassificationbehavioralanalysis}. Their framework categorizes citations as: \textit{Background} (providing context), \textit{Uses} (employing methods or tools), \textit{Compare\_contrast} (expressing similarities or differences), \textit{Motivation} (providing inspiration), \textit{Extension} (building upon methods), and \textit{Future} (suggesting future directions) as stated in Table \ref{tab:citation_datasets}. This scheme balances granularity with practical annotation feasibility (i.e., it provides a suitable compromise between classification detail and the practical ability to consistently annotate citations by hand).

Several prominent datasets have emerged using variations of these classification schemes. The SciCite dataset \citep{cohan2019structuralscaffoldscitationintent} provides 11,020 annotations using a simplified three-class scheme. The ACT dataset \citep{10.1145/3383583.3398617} contains 11,233 citations annotated by paper authors themselves, based on the assumption that authors best understand their citation intentions. The ACT2 dataset \citep{nambanoor-kunnath-etal-2022-act2} builds upon ACT with enriched contextual features while maintaining the same six-class annotation scheme.

Citation schemes are compared in Table \ref{tab:citation_datasets}. A notable characteristic of these datasets is the datasets size, which remains modest, typically in the range of a few thousand instances. This limited scale may negatively impact model training.
While Table~\ref{tab:citation_datasets} highlights the foundational datasets that shaped the field, for a more comprehensive overview of additional citation schemes and broader bibliographic databases, we refer readers to additional survey literature by \citet{jiang2023contextualised}.
\begin{table}
  \centering
  \small
  \renewcommand{\arraystretch}{1.2}
  \begin{tabular}{p{0.27\columnwidth}|p{0.10\columnwidth}|p{0.48\columnwidth}}
    \hline
    \textbf{Dataset} & \textbf{Citat.} & \textbf{Class Distribution (in \%)} \\
    \hline
    \citep{teufel2006automatic} & 2,829 & Neutral (62.7), Uses (15.8), Comp. Goals (3.9), Similar (3.8), Weakness (3.1), Others (10.7) \\
    \hline
    \citep{abu-jbara-etal-2013-purpose} & 3,500 & Criticizing, Comparison, Use, Substantiating, Basis, Neutral (Other) \\
    \hline
    \citep{10.1162/tacl_a_00028} (ACL-ARC) & 1,941 & Background (51.8), Uses (18.5), Compare/Contrast (17.5), Motivation (5.0), Extension (3.7), Future (3.5) \\
    \hline
    \citep{cohan2019structuralscaffoldscitationintent}  (SciCite) & 11,020 & Background (58), Method (29), Result Comparison (13) \\
    \hline
    \citep{10.1145/3383583.3398617}  (ACT) & 11,233 & Background (54.6), Uses (15.5), Compare\_Contrast (12.1), Motivation (9.9), Extension (6.2), Future (1.7) \\
    \hline
    \citep{nambanoor-kunnath-etal-2022-act2} (ACT2) & 4,000 & Background, Uses, Compare, Motivation, Extension, Future \\
    \hline
  \end{tabular}
  \caption{Overview of citation classification datasets and their class distributions.}
  \label{tab:citation_datasets}
\end{table}

\subsection{Computational Approaches for Citation Classification}

Early computational approaches to citation classification relied heavily on manual feature engineering, incorporating linguistic cues, citation context, and bibliometric features. Traditional machine learning rule-based methods including Support Vector Machines, Random Forests, and Naive Bayes classifiers used in studies from \citet{teufel2006automatic} or \citet{dong-schafer-2011-ensemble} dominated the field for many years.

The advent of deep learning transformed citation classification capabilities. \citet{ZHANG2025101608} provide a comprehensive survey identifying the progression from traditional machine learning to sophisticated neural architectures. Their analysis reveals increasing adoption of deep learning methods, particularly transformer-based models as BERT or SciBERT \citep{beltagy2019scibertpretrainedlanguagemodel}, for citation analysis tasks \citep{doi:10.1177/0165551521991022,nambanoor-kunnath-etal-2022-dynamic,maheshwari-etal-2021-scibert,  9319154, zhang2021tdm, zhao-etal-2019-context}.

However, as noted by \citet{ ZHANG2025101608}, no previous study has, to our knowledge, undertaken such a comparison of LLM architectures for citation function classification. The emergence of LLMs presents new opportunities for citation analysis. Nevertheless, systematic evaluation of modern LLMs for citation classification remains limited, motivating our comprehensive comparison study.

\subsection{Citation Classification Challenges}
Citation function classification faces several inherent challenges that impact model performance and evaluation. Class imbalance represents a persistent issue \citep{lyu2021}, with certain citation types (such as \textit{Compliment} or \textit{Contradiction}) appearing much less frequently than others (\textit{Background} or \textit{Uses}).  The broadest category (e.g., \textit{Background} or \textit{Neutral}) often dominates the label distribution with size over half of all citations within dataset. This imbalance can lead to models that perform well on majority classes while struggling with rare but scientifically important citation types.

In addition to label distribution issues, another critical factor influencing model performance is the scope of textual context provided to the classifier. Context extraction strategies significantly impact classification performance. Different models respond differently to varying amounts of surrounding text, from single sentences to entire paragraphs as is shown in study of \citet{nambanoor-kunnath-etal-2022-dynamic}. The optimal context window often depends on both the citation type and the underlying model architecture (see results in Section \ref{lab:results}).

\section{Methodology}

To address the limitations of existing citation classification approaches and evaluate the potential of LLMs for this task, we developed a novel dataset and conducted comprehensive experiments across multiple models and datasets. This study introduces the ArXiv Citation Classification Corpus (AC$^3$), a new dataset designed to capture nuanced citation relationships, and evaluates pre-trained language models.

We conducted our evaluation on two datasets with distinct characteristics and annotation schemes. The first dataset, ACL-ARC, provides a well-established benchmark for citation function classification with 1,941 manually annotated citations already discussed in Section \ref{subsec:citation_classification_schemes_and_datasets}. The second dataset, AC$^3$, represents our novel contribution designed to capture citation relationships that express stronger opinions and carry higher informational value. Our experiments test pre-trained language models across both datasets to assess their generalization capabilities and performance on different annotation schemes. Fine-tuning is conducted only on the ACL-ARC dataset, as the AC$^3$ dataset is too small to support additional model training. Therefore only zero-shot and few-shot approaches were tested.

\subsection{Selection of a Citation Classification Scheme}
\label{subsec:Selection_of_cit_class_scheme}

The development of AC$^3$ required careful consideration of citation classification schemes that could capture the full spectrum of citation relationships in scientific literature. Existing schemes, while comprehensive, often merge distinct citation types that carry significantly different weights in scientific discourse. For instance, the \textit{Compare\_contrast} category used by \citet{10.1162/tacl_a_00028, 10.1145/3383583.3398617} and \citet{nambanoor-kunnath-etal-2022-act2} encompasses citations that compliment other work, those that criticize it, and those that present contradictory findings, despite these having markedly different implications for scientific knowledge development.

Our annotation scheme was developed based on the work of \citet{liu2022deepgraphlearninganomalous} and other studies of citation taxonomies \citep{abu-jbara-etal-2013-purpose, 6bc982f19b0a4889993a9f34bee26ba8}, with specific extensions to better capture opinion-oriented citation relationships. The rationale for this approach centers on the observation that citations expressing strong opinions -- whether supportive or critical -- carry substantially higher informational value than neutral citations used merely for background or comparison purposes.


\begin{table}
  \centering
  \small
  \renewcommand{\arraystretch}{1.2}
  \begin{tabular}{p{0.25\columnwidth}|p{0.7\columnwidth}}
    \hline
    \textbf{Category} & \textbf{Description} \\
    \hline
    Criticizing & The citing paper describes weaknesses of the reference. \\
    Compliment & The citing paper describes strengths of the reference. \\
    Comparison & The citing paper compares present approach or results with the reference. \\
    Use & The citing paper uses the method, idea or tool of the reference. \\
    Substantiating & Current results, claims of the citing work substantiate or verify the reference or they support each other. \\
    Contradiction & Experiments with the results in the opposite way from the citing paper. \\
    Basis & The reference motivates the work of citing paper (often foundational information is provided). \\
    \hline
  \end{tabular}
  \caption{AC$^3$ annotation scheme for citation classification.}
  \label{tab:ac3_scheme}
\end{table}

Table~\ref{tab:ac3_scheme} presents our seven-category annotation scheme. This scheme distinguishes between citations that merely acknowledge prior work (\textit{Basis, Use}) and those that take explicit stances (\textit{Criticizing, Compliment, Contradiction, Comparison}). Such distinction, in our opinion, enables more nuanced analysis of scientific discourse and better captures the evaluative aspects of citation behavior. Therefore, putting them all in one category such as \textit{Compare\_contrast} as proposed in other papers like \citet{10.1162/tacl_a_00028, nambanoor-kunnath-etal-2022-act2} or  \citet{10.1145/3383583.3398617}, diminishes their citation value.

To prevent conceptual ambiguity among supportive relationships -- specifically distinguishing the objective empirical verification in \textit{Substantiating} from the subjective praise in \textit{Compliment} -- we carefully delineate the boundaries of these classes. To further clarify these distinctions, concrete textual examples for each category are provided in Appendix~\ref{sec:appendix_examples}.

\subsection{Annotating Citations}

The AC$^3$ dataset was constructed through manual annotation of citations extracted from ArXiv publications across multiple scientific domains. Unlike many existing citation-function datasets that are predominantly NLP-focused, our dataset deliberately encompasses a diverse range of disciplines to better assess the generalizability of citation behavior. Specifically, we collected papers from nine distinct ArXiv categories covering following scientific domains: Condensed Matter (\texttt{cond-mat}), Computation and Language (\texttt{cmp-lg}), Astrophysics (\texttt{astro-ph}), Atomic Physics (\texttt{atom-ph}), Algebraic Geometry (\texttt{alg-geom}), Accelerator Physics (\texttt{acc-phys}), Atmospheric-Oceanic Sciences (\texttt{ao-sci}), Adaptation and Self-Organizing Systems (\texttt{adap-org}), and Cellular Automata and Lattice Gases (\texttt{comp-gas}).\footnote{Information about scientific domain is included in the dataset explicitly.}

Unlike datasets created through crowd-sourcing or expert annotation by linguistic specialists, our annotations were performed by the research team, ensuring consistency in interpretation while acknowledging the inherent subjectivity in determining citation intent.

The dataset construction process involved several key steps. Publications were selected from various ArXiv subject categories to ensure comprehensive representation of citation formats and scientific domains. Each selected publication was manually reviewed to identify citations appearing in multiple formats, including numerical references (\textit{[15]}), grouped citations (\textit{[3,4,5] [9]}), range citations (\textit{[9]-[11}]), author-year citations (\textit{Hasselmann, 1971}), and mixed notation systems (\textit{10)}; \textit{[N3], [N7], [N8]}...). This approach ensured basic coverage of citation styles commonly found in scientific literature.

The annotation was performed by two experienced annotators as follows. Before starting, the annotators created detailed guidelines, taking inspiration from the ACL-ARC~\citep{10.1162/tacl_a_00028} and ACT2~\citep{nambanoor-kunnath-etal-2022-act2} datasets. Annotation was carried out iteratively: after completing a small portion of the samples, the annotators reviewed their work, discussed ambiguities, and updated the guidelines accordingly. This iterative strategy helped achieve high inter-annotator agreement and addressed most issues early in the process.

We evaluated inter-annotator agreement using the Cohen’s Kappa metric~\citep{cohen1960coefficient}. After annotating the first data sample, the initial agreement was 52\%. Following the iterative refinement of the guidelines and discussion of annotation issues, the final Cohen’s Kappa across the full dataset increased to 71\%.

\begin{table}
  \centering
      \renewcommand{\arraystretch}{1.2}
  \begin{tabular}{lc}
    \hline
    \textbf{Category} & \textbf{Citation Count} \\
    \hline
    Basis & 159 \\
    Substantiating & 148 \\
    Use & 133 \\
    Comparison & 46 \\
    Compliment & 25 \\
    Criticizing & 9 \\
    Contradiction & 7 \\
    \hline
    \textbf{Total} & \textbf{530} \\
    \hline
  \end{tabular}
  \caption{Class distribution in the AC$^3$ dataset.}
  \label{tab:ac3_distribution}
\end{table}

Table~\ref{tab:ac3_distribution} presents the class distribution within AC$^3$. The data reveals significant class imbalance, with neutral
citation types (\textit{Basis, Substantiating, Use}) heavily dominating the dataset, while opinion-oriented citations (\textit{Criticizing, Compliment, Contradiction}) remain severely underrepresented. This distribution reflects the natural scarcity of explicit evaluative citations in the scientific literature.
The class imbalance simultaneously presents both opportunities and challenges for model development. While the underrepresentation of rare but valuable critical and contradictory citations reflects realistic scientific writing patterns, it also poses significant challenges for training LLMs to detect these important but infrequent citation types reliably.

\subsection{Context Extraction Method}

Citation context extraction represents a critical component of citation function classification, as the surrounding text provides essential cues for determining citation intent. Unlike fixed context windows used in many existing approaches, we implemented a  context extraction method in different ways.

Our context extraction strategy was implemented in four distinct variants to systematically evaluate the impact of context scope on classification performance. The first variant extracts the full paragraph containing the citation, providing maximum contextual information but potentially introducing noise from unrelated content. The second variant uses a focused window of 30 words before and 10 words after the citation. The third and fourth variants extract two and three sentences surrounding the citation.

The choice of context extraction strategy proves crucial for model performance, as different models respond differently to varying amounts of surrounding text \citep{nambanoor-kunnath-etal-2022-dynamic}. This systematic approach enables optimization of context windows for specific model architectures and citation types, rather than applying a universal context extraction method that may be suboptimal for certain model-citation combinations.

\subsection{Datasets Split}

For the ACL-ARC dataset, we split the 1,941 citations into 87\% for training, another roughly 7\% for testing, and roughly 6\% for validation. For comparison with other studies we used same division as \citet{cohan2019structuralscaffoldscitationintent} (see Section \ref{subsubsec:comp_sota}).
The AC$^3$ dataset was used solely for testing pre-trained language models due to its limited size, therefore, no dataset split was required.

\subsection{Language Models}
The models evaluated in this study include five SOTA language models: Mistral 7B \citep{jiang2023mistral7b}, Orca 2-7B \citep{mitra2023orca2teachingsmall}, LLaMA 3.1-8B \citep{dubey2024llama3herdmodels}, Falcon 7B \citep{almazrouei2023falconseriesopenlanguage} and SciBERT \citep{beltagy2019scibertpretrainedlanguagemodel}. For testing ACL-ARC dataset, we fine-tuned all models using Hugging Face Transformers library\footnote{https://github.com/huggingface/transformers} \citep{wolf-etal-2020-transformers}.

\subsection{Experimental Setup}
We evaluate models across three paradigms: zero-shot prompting without task-specific examples, few-shot prompting using one example per citation class, and fine-tuning with full supervised training on the training dataset. For the fine-tuning paradigm, rather than employing a text generation approach, we formulate the task as sequence classification. Specifically, we attach a linear classification head on top of the pooled output from the model (utilizing last-token pooling for the causal language models and the standard \texttt{[CLS]} token representation for SciBERT). 
All models utilize the AdamW optimizer \citep{loshchilov2019decoupledweightdecayregularization} with model-specific learning rates determined through systematic experimentation. Mistral 7B and Orca 2-7B employ learning rates of 5e-6, LLaMA 3.1-8B uses 2e-6, Falcon 7B operates at 5e-5, and SciBERT requires a learning rate of 2e-5.

Learning rate scheduling employs linear decay with warmup periods of 10-15\% of total training steps (0\% for SciBERT). Batch sizes vary based on available GPU memory constraints, with larger models using smaller batch sizes. LLaMA 3.1-8B uses 8-bit quantization while Falcon 7B implements 4-bit quantization for memory optimization.

Class imbalance is addressed through weighted cross-entropy loss function. Early stopping monitors validation performance using either F1 macro score or validation loss depending on the model. All experiments are implemented in PyTorch with fixed random seeds for reproducibility. LLaMA 3.1-8B and Falcon 7B were additionally trained with reduced learning rates.

\subsubsection{Evaluation Metrics}
Macro F1 score has emerged as the preferred evaluation metric for citation classification, as it provides equal weight to all classes regardless of their frequency. This metric better reflects model performance on minority classes compared to accuracy or weighted metrics that can be dominated by majority class performance.

\section{Results}
This section presents the obtained results.
\subsection{Citation Classification Performance on ACL-ARC Dataset}
Table~\ref{tab:results_acl_arc_part} presents comprehensive results for all models across the three evaluation paradigms on the ACL-ARC dataset. Our experiments reveal substantial performance variations both across models and training approaches. Table~\ref{tab:results_acl_arc_whole} presents the results of only pre-trained models on the whole dataset. The comparison indicates that classification on the test split appears easier, as overall performance decreases when evaluated on the full dataset. 

\begin{table}
  \centering
  \small
    \renewcommand{\arraystretch}{1.2}
    \begin{tabular}{p{0.20\columnwidth}|>{\centering\arraybackslash}m{0.19\columnwidth}|
                  >{\centering\arraybackslash}m{0.19\columnwidth}|
                  >{\centering\arraybackslash}m{0.22\columnwidth}}
    \hline
    \textbf{LLM} & \textbf{Zero-shot} & \textbf{Few-shot} & \textbf{Fine-tuning} \\
    \hline
    Mistral 7B & 29.07 & \textbf{49.68} & 67.55 \\
    Orca 2-7B & \textbf{47.63}  & 44.89 & 68.31 \\
    LLaMA 3.1-8B & 33.13  & 33.05 & 61.66  \\
    Falcon 7B & 1.38 & 3.38 & \textbf{76.35} \\
    SciBERT & - & - & 75.14  \\
    \hline
  \end{tabular}
  \caption{Performance comparison on ACL-ARC dataset (139 citations) [in \%].}
  \label{tab:results_acl_arc_part}
\end{table}

\begin{table}
  \centering
  \small
    \renewcommand{\arraystretch}{1.2}
  \begin{tabular}{l|c|c}
    \hline
    \textbf{LLM} & \textbf{Zero-shot} & \textbf{Few-shot}  \\
    \hline
    Mistral 7B & 23.11 & \textbf{41.13}  \\
    Orca 2-7B & \textbf{33.13} & 34.14 \\
    LLaMA 3.1-8B & 24.71 & 24.68 \\
    Falcon 7B & 1.46 & 4.82  \\
    \hline
  \end{tabular}
  \caption{Performance comparison of selected pre-trained LLMs using zero-shot and few-shot across whole dataset ACL-ARC (1,941 citations) [in \%].}
  \label{tab:results_acl_arc_whole}
\end{table}

The fine-tuning experiments reveal that Falcon 7B achieved the highest performance among the tested models, reaching 76.35\% macro F1 score. 

Notably, Falcon 7B shows the most dramatic improvement from fine-tuning despite poor zero-shot and few-shot performance. This pattern suggests that some models may require task-specific optimization to effectively leverage their capabilities for citation classification. Interestingly, although Falcon 7B does assign citations to different categories -- unlike SciBERT, which is not included in the zero- or few-shot tests -- it still performs very poorly without fine-tuning. This indicates that Falcon likely does not understand the task well in its pre-trained form and needs extensive adaptation to perform competitively.

Orca 2-7B demonstrates the strongest zero-shot performance (47.63\% or 33.13 on whole dataset), indicating superior out-of-the-box capabilities for this task. However, this advantage diminishes with fine-tuning, where other models achieve comparable or superior performance. This strong zero-shot performance may be partially attributed to design of Orca~2 as a reasoning-oriented model \citep{mitra2023orca2teachingsmall}.

SciBERT, despite its smaller size and domain-specific training, achieves competitive results (75.14\% macro F1), demonstrating the value of scientific domain adaptation for citation classification tasks (see Table \ref{tab:results_acl_arc_part}).

\subsubsection{Comparison with SOTA}
\label{subsubsec:comp_sota}
The evaluation on this dataset is usually performed on different data splits.
Only~\citet{cohan2019structuralscaffoldscitationintent}\footnote{\url{https://github.com/allenai/scicite}} provide the exact details of the dataset split they used: 1,688 instances for training, 114 for validation, and 139 for testing. 
\citet{nambanoor-kunnath-etal-2022-dynamic} and \citet{10.1145/3583780.3615018} follow the train (1,647 instances) and test split (284 instances) with 1931 instances. \citet{10.1162/tacl_a_00028}
combine two dataset (ACL-ARC with \citet{teufel2006automatic}). The resulting citation function dataset contains 3,083 instance.

Therefore, to enable a fair comparison with related work, we selected the best-performing model (the fine-tuned Falcon) and adopted the same split strategy as suggested by \citet{cohan2019structuralscaffoldscitationintent}.

Table~\ref{tab:sota_comparison} presents the comparison of results with previous related studies, highlighting the best performance achieved by the Falcon 7B LLM in our experiments.
This table clearly demonstrates that our best model outperforms the others by a significant margin.

\begin{table}
    \centering
    \small
    \renewcommand{\arraystretch}{1.2}
    \begin{tabular}{p{0.6\columnwidth}|p{0.25\columnwidth}}
                    \hline
    \textbf{Approach} & \textbf{F1 Macro}  \\ \hline
\citep{10.1162/tacl_a_00028}  &  53.0\\     \hline
\citep{cohan2019structuralscaffoldscitationintent} & 67.9 \\     \hline
\citep{nambanoor-kunnath-etal-2022-dynamic}  &  70.8\\     \hline
\citep{10.1145/3583780.3615018}  &  64,5\\     \hline
 \citep{10.1109/JCDL57899.2023.00017} & 68.4 \\ \hline 
 \hline
 
Fine-tuned Falcon 7B (proposed)  &  \textbf{73.3}\\     \hline

    \end{tabular}
    \caption{Performance comparison on ACL-ARC with SOTA results [in \%].}
    \label{tab:sota_comparison}
\end{table}

\subsection{Error Analysis}
The analysis of most errors involves the \textit{Background} class and this class is sometimes confused with \textit{Compare/Contrast} and \textit{Uses}. \textit{Background} is misclassified most often, likely because it is the largest class in the dataset and its general contextual nature makes it overlap with more specific citation functions. Both \textit{Compare/Contrast} and \textit{Uses} also occasionally misclassify into each other, likely because citations often both compare different approaches and describe how they are used.

The \textit{Motivation} class shows the weakest performance and is frequently confused with \textit{Background}, likely because motivational statements often provide general context about why research is needed, making them difficult to distinguish from general background information.

\subsection{Citation Classification Performance on AC$^3$ Dataset}
\label{lab:results}
A critical aspect of our evaluation involves examining how different citation context extraction strategies impact classification performance across different models.
To evaluate this impact, we tested all mentioned pre-trained models (except SciBERT) across four variants of our AC$^3$ dataset, each employing different citation context extraction methods. Table~\ref{tab:ac3_results} presents the zero-shot and few-shot performance across these variants. The best results in each method and model are highlighted. 

\begin{table}
  \centering
  \small
\renewcommand{\arraystretch}{1.2}
  \begin{tabular}{c|l|c|c}
    \hline
    \textbf{Version} & \textbf{LLM} & \textbf{Zero-shot} & \textbf{Few-shot} \\
    \hline
    1 & Mistral 7B & 8.79 & 15.13 \\
    2 & Mistral 7B & \textbf{16.74} & \textbf{22.43} \\
    3 & Mistral 7B & 15.75 & 17.29 \\
    4 & Mistral 7B & 16.01 & 18.18 \\
    \hline
    1 & Orca 2-7B & 16.97 & 10.90 \\
    2 & Orca 2-7B & \textbf{20.90} & \textbf{18.76} \\
    3 & Orca 2-7B & 20.06 & 16.75 \\
    4 & Orca 2-7B & 19.78 & 11.57 \\
    \hline
    1 & LLaMA 3.1-8B & 10.78 & \textbf{17.41} \\
    2 & LLaMA 3.1-8B & 11.75 & 14.21 \\
    3 & LLaMA 3.1-8B & 12.49 & 14.00 \\
    4 & LLaMA 3.1-8B & \textbf{12.74} & 16.31 \\
    \hline
    1 & Falcon 7B & 6.35 & \textbf{9.29} \\
    2 & Falcon 7B & \textbf{7.04} & 8.94 \\
    3 & Falcon 7B & 5.74 & 7.52 \\
    4 & Falcon 7B & 5.62 & 8.32 \\
    \hline
  \end{tabular}
  \caption{AC$^3$ dataset performance (macro F1 \%) across context variants. Version 1: Full paragraph in which sentence occurs.; Version 2: Thirty words preceding the citation and ten words following it, all within the same paragraph. If the paragraph is shorter, the citation content contains less words.; Version 3: Two sentences (each containing at least three words), with the final sentence containing the reference. If there is no preceding sentence within the paragraph, only the sentence with the citation is used.; Version 4: Three sentences, where the middle one is containing the reference. If the paragraphs is shorter, the citation content contains less words or sentences.}
  \label{tab:ac3_results}
\end{table}

The results demonstrate that optimal context extraction strategies vary significantly across models. Mistral 7B achieves best performance with Version 2 (30 words before, 10 after citation) in both zero-shot (16.74\%) and few-shot (22.43\%) settings. Orca 2-7B also shows the strongest zero-shot results when used with Version 2 (20.90\%), though its few-shot performance when providing examples is much worse. LLaMA 3.1-8B shows relatively consistent performance across context variants, with slight preferences for longer contexts. Falcon 7B demonstrates the most consistent performance across variants, though at substantially lower absolute scores than other models.
The performance variations across context extraction methods highlight the importance of model-specific optimization for citation classification tasks. The finding that Version 1 (full paragraph) generally yields poorer results suggests that excessive context may introduce noise that impairs classification accuracy.

\subsection{Comparative Analysis}

The substantial performance gap between zero-shot/few-shot and fine-tuning approaches highlights the complexity of citation function classification. Even SOTA LLMs struggle to reliably classify citations without task-specific training, suggesting that this task requires specialized knowledge beyond general language understanding.

The dramatic improvement observed across all models (20-70 percentage point increases in macro F1) underscores the importance of supervised learning for this domain.

Model size does not directly correlate with performance, as evidenced by Falcon 7B outperforming the larger LLaMA 3.1-8B model. This suggests that model architecture or training data quality of LLMs may be more important than parameter count for citation classification tasks.

\subsection{Discussion}

\begin{figure}[t]
  \includegraphics[width=\columnwidth]{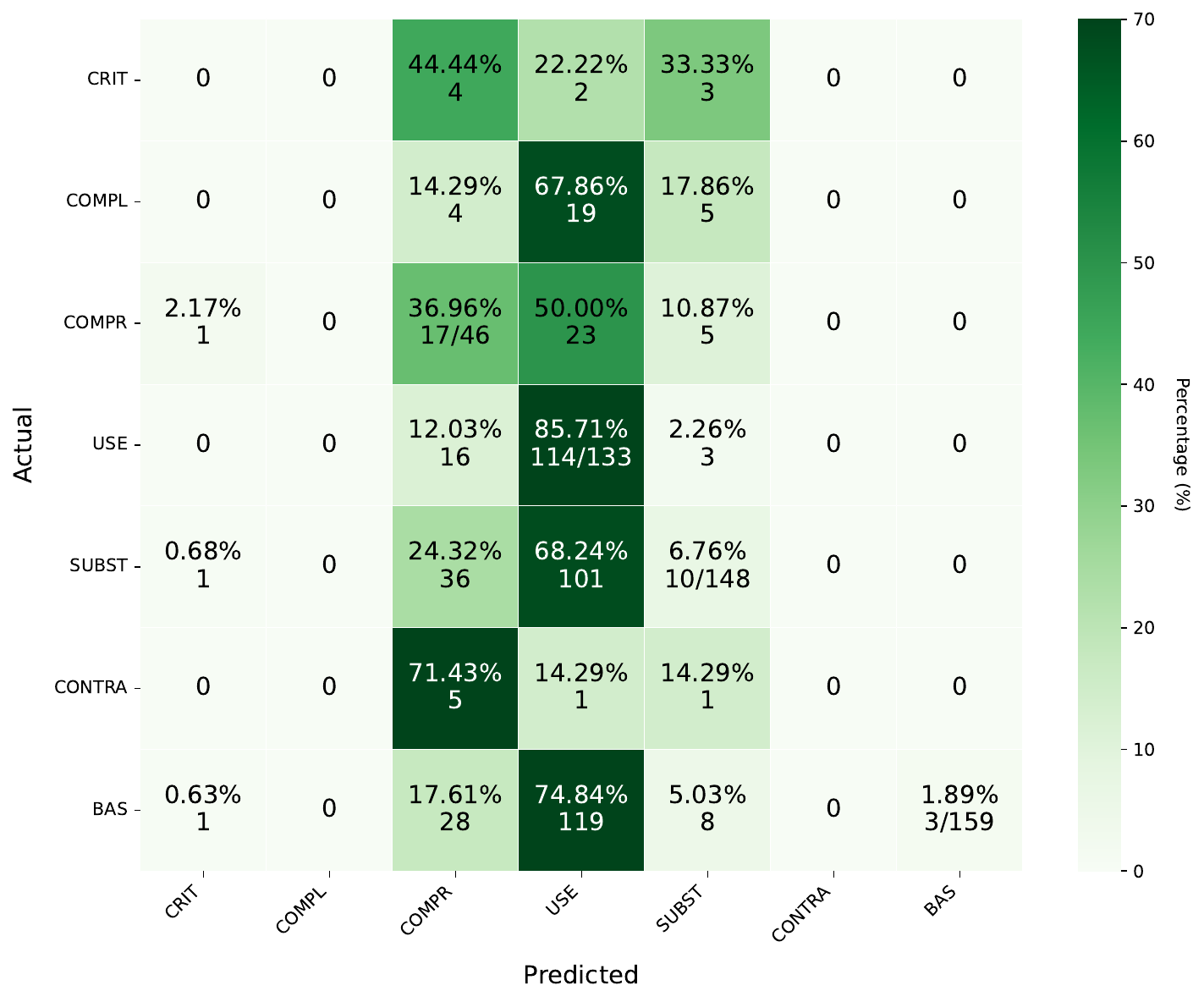}
  \caption{Confusion matrix for Orca 2-7B classification using few-shot on the AC\textsuperscript{3} dataset -- version with three sentences.}
  \label{fig:experiments}
\end{figure}


Our results reveal several important insights about LLM performance for citation classification. Figure~\ref{fig:experiments} presents the~confusion matrix for citation classification using few-shot for Orca-7B model (AC$^3$ dataset). It is evident that this model struggles with classifying into the~categories \textit{Criticizing}, \textit{Compliment} and \textit{Contradiction}. Although these classes are infrequent in~the~dataset, they appear even less frequently (or not at all) in the~model's predictions. This pattern is consistent across all tested models using few-shot (to some extent it is also visible during zero-shot). Unfortunately, these categories hold the~highest significance, as they convey particularly strong opinions. This phenomenon is~particularly distinguishable in~the~case of~Orca model and it~can be~explained by~several factors.

As stated in Section \ref{subsec:Selection_of_cit_class_scheme} these categories may require a~deeper understanding of author intent and subtle sentiment analysis, which can be challenging even for~the~LLMs. The fact that these categories "convey particularly strong opinions" suggests they have more complex linguistic patterns that require sophisticated reasoning, which models like the~Orca 2-7B might struggle to capture without the~sufficient training examples. 

Furthermore, the~few-shot prompting approaches may not provide enough contextual guidance for~the~models to recognize these rare but semantically complex citation types. 
To address this issue, fine-tuning proves essential for achieving competitive performance.

Class imbalance remains still a persistent challenge across all fine-tuning models. Although weighted loss functions help reduce this issue, fine-tuned LLMs still struggle with minority classes that may be crucial for comprehensive citation analysis.

\section{Conclusions}


Our work addresses a significant gap identified in recent surveys by providing the first dedicated comparison of LLM architectures for citation classification. 

This study presents on of the first comprehensive evaluations of modern LLMs for citation function classification, achieving new SOTA results on the ACL-ARC dataset. Our fine-tuned Falcon 7B model reaches 73,3\% macro F1 score, representing substantial improvement over previous approaches.


Beyond achieving superior performance on existing dataset, this work introduces AC$^3$, a new citation dataset that captures nuanced citation relationships that express stronger opinions and carry higher informational value than traditional citation categories. It uses a refined seven-category annotation scheme to distinguish between neutral references and explicit evaluative stances, enabling deeper analysis of citation behavior. The dataset includes four context extraction variants to assess how contextual scope affects LLM classification performance.

Key findings from our comparison study include: (1) Fine-tuning proves essential for competitive performance, with dramatic improvements over zero-shot and few-shot prompting approaches; (2) Model size does not directly correlate with citation classification performance; (3) We assume that, domain-specific models (tested only on SciBERT) remain competitive despite smaller parameter counts (110M vs 4-8B parameters); (4) Pre-trained models exhibit significant difficulty with classification into rare citation classes, despite these categories carrying substantial weight in scientific discourse; (5) The novel dataset experiments revealed that optimal citation context extraction strategies are model-dependent rather than universal, challenging the assumption that a single approach to context extraction would be optimal across different model architectures.

\section{Limitations}

Several limitations constrain our study's scope and generalizability. 

The class imbalance inherent in citation datasets, particularly pronounced in our AC$^3$ corpus where opinion-oriented citations represent less than 8\% of all instances, affects all models tested, potentially limiting performance on rare but important citation types. While we implemented strategies like weighted loss function, the fundamental challenge of detecting infrequent but semantically complex citation categories remains difficult to address completely without substantially larger datasets or specialized training techniques.

The optimal balance between model performance and computational efficiency requires further investigation.

Our partially dynamic context extraction strategy with predetermined window sizes and sentence boundaries may not be optimal for citation function classification. Truly dynamic context selection could potentially improve performance but introduces additional complexity.


Finally, our analysis focuses on English-language scientific literature. Extension to other languages would broaden the applicability of our findings.

\section*{Acknowledgements}
This work has been partly supported by the Grant No. SGS-2025-022 -- New Data Processing Methods in Current Areas of Computer Science.
Computational resources were provided by the e-INFRA CZ project (ID:90254), supported by the Ministry of Education, Youth and Sports of the Czech Republic.

\section{Bibliographical References}\label{sec:reference}
\bibliographystyle{lrec2026-natbib}
\bibliography{lrec2026-example}

\appendix

\section{Citation Category Examples}
\label{sec:appendix_examples}

Table~\ref{tab:ac3_examples} provides concrete examples for each of the seven citation categories to further clarify their distinctions and practical boundaries.

\begin{table}[h!]
  \centering
  \small
  \renewcommand{\arraystretch}{1.4} 
  \begin{tabular}{p{0.25\columnwidth}|p{0.7\columnwidth}}
    \hline
    \textbf{Category} & \textbf{Example of a Citation Sentence} \\
    \hline
    Criticizing & ``However, the approach proposed by Smith et al. (2020) suffers from high computational complexity and fails to scale on larger datasets.'' \\
    Compliment & ``Jones et al. (2021) introduced an elegant and highly robust framework that remains the gold standard for text classification.'' \\
    Comparison & ``Our proposed model achieves a 5\% higher F1-score on the benchmark dataset compared to the baseline established by Davis et al. (2019).'' \\
    Use & ``For the initial data preprocessing and tokenization steps, we utilized the open-source pipeline provided by Lee et al. (2022).'' \\
    Substantiating & ``Consistent with the early observations of Brown et al. (2018), our empirical results confirm that increasing the dropout rate prevents overfitting in this specific architecture.'' \\
    Contradiction & ``In contrast to White et al. (2020), who reported a positive correlation, our extensive analysis reveals a slight negative correlation between these two variables.'' \\
    Basis & ``The theoretical foundation for dynamic graph embeddings was formally established in the seminal work of Clark et al. (2017).'' \\
    \hline
  \end{tabular}
  \caption{Concrete textual examples for each category in the AC$^3$ annotation scheme, illustrating the practical boundaries between distinct citation relationships.}
  \label{tab:ac3_examples}
\end{table}

\end{document}